\documentclass[final]{cvpr}

\usepackage{times}
\usepackage{epsfig}

\usepackage{array,amssymb,amsthm,amsmath,amstext,booktabs,bm,enumerate,color,cite,float,graphicx,soul,multirow,threeparttable,subfigure}
\usepackage{algorithm}
\usepackage{algorithmic}

\usepackage[pagebackref=true,breaklinks=true,colorlinks,bookmarks=false]{hyperref}

\def\cvprPaperID{4096} 
\def\confYear{CVPR 2021}

\begin{document}

\title{PICA: A Pixel Correlation-based Attentional Black-box Adversarial Attack}

\author{Jie Wang, Zhaoxia Yin, Jin Tang, Jing Jiang, and Bin Luo\\

}

\maketitle

\begin{abstract}
The studies on black-box adversarial attacks have become increasingly prevalent due to the intractable acquisition of the structural knowledge of deep neural networks (DNNs). However, the performance of emerging attacks is negatively impacted when fooling DNNs tailored for high-resolution images. One of the explanations is that these methods usually focus on attacking the entire image, regardless of its spatial semantic information, and thereby encounter the notorious curse of dimensionality. To this end, we propose a pixel correlation-based attentional black-box adversarial attack, termed as PICA. Firstly, we take only one of every two neighboring pixels in the salient region as the target by leveraging the attentional mechanism and pixel correlation of images, such that the dimension of the black-box attack reduces. After that, a general multiobjective evolutionary algorithm is employed to traverse the reduced pixels and generate perturbations that are imperceptible by the human vision. Extensive experimental results have verified the effectiveness of the proposed PICA on the ImageNet dataset. More importantly, PICA is computationally more efficient to generate high-resolution adversarial examples compared with the existing black-box attacks.

\end{abstract}

\begin{figure}[htbp]
	\centering
	\subfigure[Adversarial example generated by the baseline \cite{suzuki2019adversarial}]{
		\label{1a}
		\includegraphics[width=3.2in]{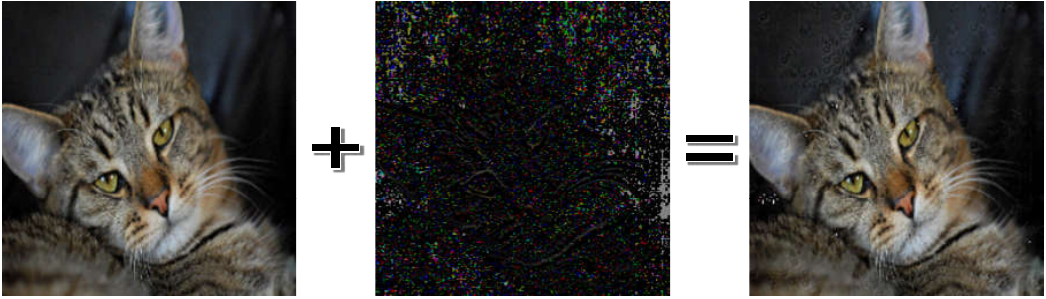}
	}
	\subfigure[Adversarial example generated by PICA]{
		\label{1b}
		\includegraphics[width=3.2in]{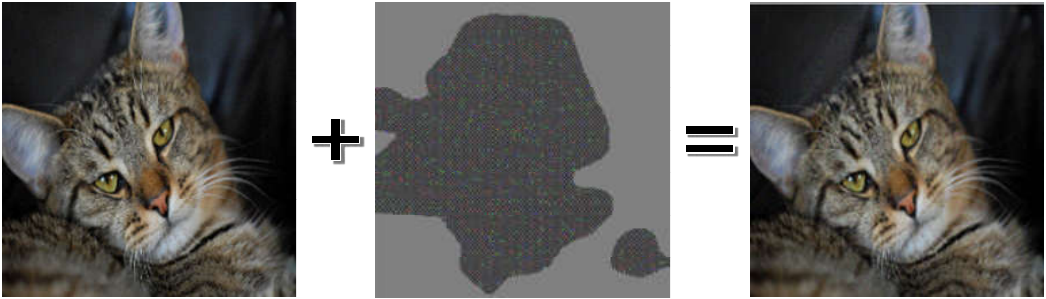}
	}
	\caption{The original images, perturbation patterns, and corresponding AEs generated by the two methods, respectively (In each perturbation pattern, a gray pixel indicates the pixel of the original image has not been modified. Besides, a brighter pixel indicates an intensity change, while a darker one represents a change in the opposite direction). }
	\label{fig:1}
\end{figure}

\section{Introduction}
Deep Neural Networks (DNNs) have been proven fruitful in many realistic applications that involve machine learning (ML) and artificial intelligence (AI) \cite{jiang2020efficient}. Despite the phenomenal success, DNNs are usually vulnerable, that is, they may yield wrong categories to adversarial examples (AEs) in pattern recognition tasks \cite{szegedy2013intriguing,goodfellow2014explaining,moosavi2016deepfool}.  AEs are firstly explained as a kind of manual images, where several invisible perturbations are imposed compared with the original ones. Apart from the image processing, AEs are also widely found in other fields, such as the natural language processing \cite{alzantot2018generating} and speech recognition \cite{alzantot2018did,zhang2020generating}. Therefore, the concern about the robustness of DNNs has been raised, stimulating a line of studies on AI safety issues in the real world \cite{kurakin2016adversarial,Eykholt_2018_CVPR}.

Since AEs reveal the weakness of DNNs, investigating the generation of AEs can help ML engineerings to employ more robust DNNs. Existing adversarial attacks tend to take advantage of the gradient-based optimization to achieve AEs under the white-box setting \cite{carlini2017towards,goodfellow2014explaining,moosavi2016deepfool,rozsa2016adversarial,madry2017towards}. Nevertheless, these methods are reasonable only if the researchers fully know the structural knowledge of the target DNN. Usually, there is no sufficient information about the neural architecture, hyperparameters, or training data. As a result, the black-box adversarial attacks have drawn much attention during the past years.

One of the representative works firstly fooled a substitute white-box DNN and transferred the generated AEs to the target black-box model \cite{dong2019evading}. This study relaxes the necessity of the inner information of DNNs but struggles to the mismatch between the substitute model and the target one. The emerging attacks also attempted to estimate the gradients of DNNs using various technologies \cite{tu2019autozoom}. However, these attacks are still computationally expensive as they often resort to many kinds of optimization techniques \cite{chen2017zoo,alzantot2019genattack}.

More recently, a few black-box attacks often work by defining an objective function and interfering with the input iteratively, hoping to obtain AEs with evolutionary algorithms (EAs) \cite{su2019one, liu2020black}. Typically, they first formulate the objective function as an aggregation of the confidence probability of the true label and the strength of perturbations with a penalty. Through evolution operators, the candidate AEs have the potential to confuse DNNs without being perceived by human vision. However, this class of attacks makes the optimization result sensitive to the value of the penalty, such that it cannot balance two objectives and inevitably miss the optimal trade-off perturbation.

Furthermore, it is worth noting that the performance of EA-based black-box attacks is negatively impacted when fooling DNNs tailored for high-resolution images (\emph{e.g.}, see Fig. \ref{1a}). One of the possible explanations is that these methods usually focus on attacking the entire image, where the perturbation on each pixel is considered as a decision variable in the optimization, and thereby encounter the notorious curse of dimensionality \cite{yang2008large}. To overcome the above drawbacks, we propose PICA, a pixel correlation-based attentional black-box adversarial attack. The main contributions of this study can be summarized as follows.
\begin{enumerate}[i.]
\item \textbf{Utilizing the attentional mechanism to screen the attacked pixels.} We firstly use the class activation mapping (CAM) and a proxy model to obtain the attentional map of the target image. The map strictly limits the attacked pixels so that the perturbations are only allowed to emerge within the salient region. On the one hand, attacking the salient pixels might be more efficient than the entire image, as these pixels can better reflect the spatial semantic information. On the other hand, screening the pixels is able to reduce the dimensionality of decision variables in the case of high-resolution images, which is beneficial to the black-box optimization.
\item \textbf{Leveraging the pixel correlation of the image to further refine the target pixels.} We secondly follow the parity-block pattern and divide the image into two segments, that is, only one of every two adjacent pixels is considered as the target. The rationality of this implementation can be explained as follows: the neighboring pixels usually share similar information and pixel values. Using the pixel correlation may avoid redundant perturbations and further reduce the decision variables of the black-box optimization.
\item \textbf{Performing the black-box adversarial attack with a multiobjective evolutionary algorithm.}  We finally resort to a multi-objective evolutionary algorithm (MOEA) to optimize the perturbations on high-resolution images. Through balancing two objectives, the MOEA will generate effective AEs that easily fool DNNs while being imperceptible by human vision (see Fig. \ref{1b}). Note that, any MOEA can be embedded into PICA without additional techniques for dealing with large-scale decision variables, which is computationally efficient.
\end{enumerate}

The remainder of the paper is structured as follows. Section II reviews the existing researches on adversarial attacks and motivation of this paper. Section III introduces the methodology of the proposed PICA. Experimental results and ablation studies are given in Section IV. Section V finally provides the conclusions and future works.

\section{Related works}
\subsection{White-box adversarial attacks}
Over the past few years, extensive researches on adversarial attacks have emerged. We can categorize them into two classes, including white-box and black-box attacks. For the former, the scholars can fully recognize the structural knowledge of the target DNN and thereby calculate gradients to realize an efficient gradient-based attack \cite{dong2018boosting}. Goodfellow et al. \cite{goodfellow2014explaining} proposed the fast gradient sign method (FGSM), which can generate AEs according to the categorical loss gradient of the input image. Following this, they also proposed an iterative version of FGSM (IFGSM) for better performance \cite{kurakin2016adversarial1}. Limited to the pages, we have to omit the other relevant references and more comprehensive review on white-box attacks can refer to \cite{zhang2019adversarial}.

\subsection{Black-box adversarial attacks}
A variety of black-box attacks take advantage of the  transferability \cite{shi2019curls,papernot2017practical} and gradient estimation to reach the goal \cite{tu2019autozoom,nitin2018practical}. Transferability-based attacks generally trained a DNN as a substitute at first. Then, they performed a white-box attack, expecting the gradient information of the substitute can help to fool the target DNN. It is clear that this type of attack relaxes the necessity of the inner information of DNNs, but heavily depends on the transferability, which is sensitive to the mismatch between the substitute and target DNN. More seriously, the computational cost of training a substitute becomes unacceptable when confronting large-scale DNNs. Other than transferability-based attacks, Chen \emph{et al.} \cite{chen2017zoo} suggested a C\&W attack, where the loss function was modified to make itself only rely on the output of DNNs. Then, C\&W estimated the gradients with finite differences and performed the white-box adversarial attack.  Although the gradient estimation-based attacks are independent of the transferability, the subsequent studies have revealed that they may not alleviate the computational cost as they often require a large number of queries \cite{alzantot2019genattack}.

\begin{figure*}
	\begin{center}
		\includegraphics[width=6.5in]{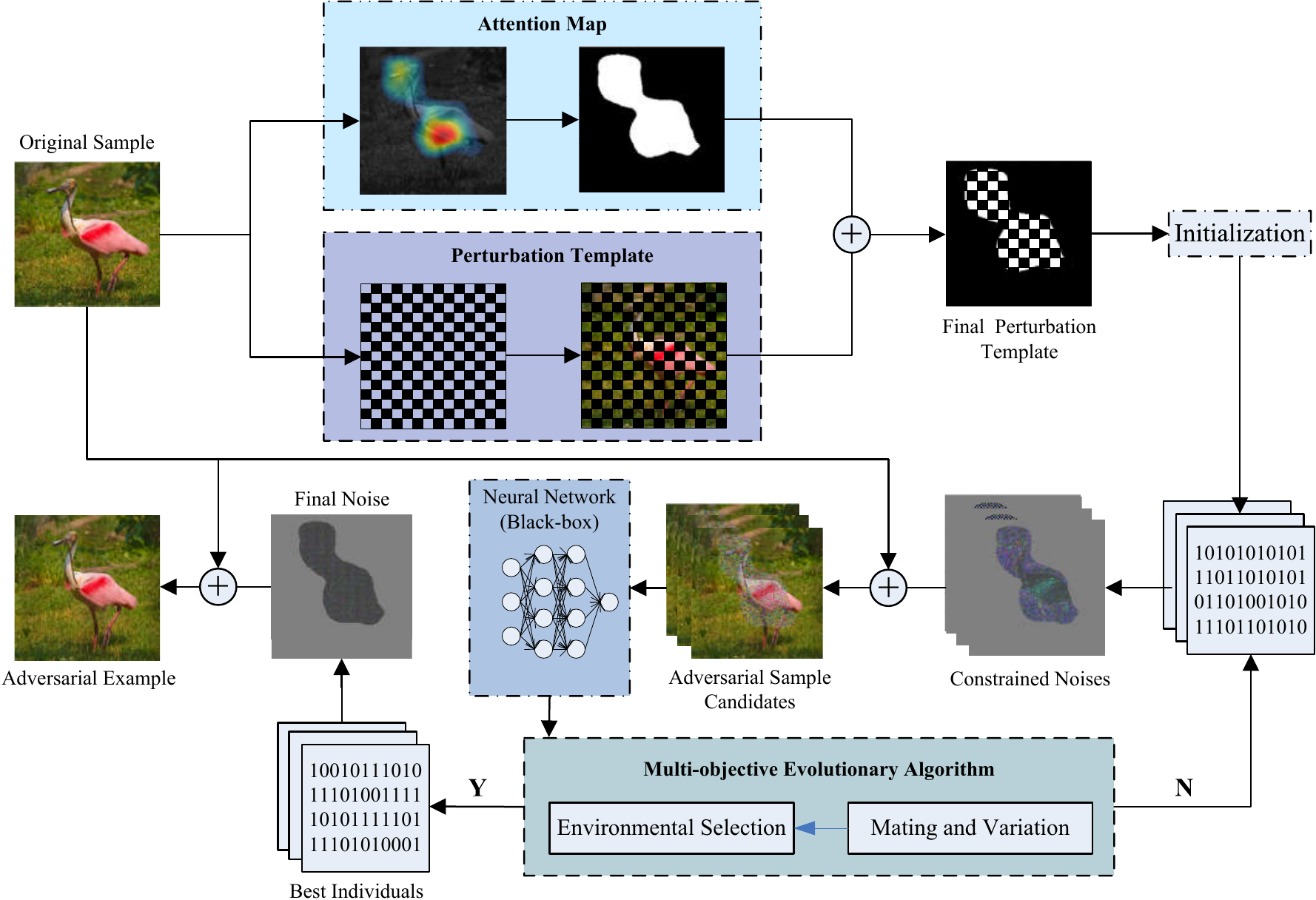}
	\end{center}
	\caption{The general framework of the proposed PICA.}
	\label{fig:framework}
\end{figure*}
\subsection{Evolutionary algorithm based attacks}
EA is competitive in solving black-box attacks by exploring the optimal AEs due to its efficient gradient-free global search \cite{fonseca1995overview,mukhopadhyay2013survey,han2019survey}. Su \emph{et al.} \cite{su2019one} proposed a one-pixel black-box attack based on differential evolution (DE) \cite{storn1997differential}, where only a few pixels or even one pixel was perturbed. Liu \emph{et al.} \cite{liu2020black} proposed another approach that focused on attacking the entire image instead of one pixel. They modeled the adversarial attack as a constrained optimization problem and optimized the penalty function via a genetic algorithm. In \cite{suzuki2019adversarial}, the scholars pointed out that the optimization result might be sensitive to the value of the penalty. Therefore, they reformulated black-box adversarial attacks as multiobjective optimization problems (MOPs) and solved them with MOEAs. To be specific, they viewed the classification probability and visual quality of AEs as two conflicting objective functions.

The above approaches are suitable for attacking low-resolution images (\emph{e.g.}, CIFAR-10). However, their performance will degenerate if the dimension of images increases (\emph{e.g.}, ImageNet). For \cite{su2019one}, attacking limited pixels might fail to influence the confidence of classification. In other words, the method cannot ensure the success rate of the attack in the case of high-resolution images.  As for \cite{suzuki2019adversarial}, attacking the entire image will result in large-scale decision variables and the curse of dimensionality, which affects the performance of MOEAs.

\subsection{Motivation}
Based on the above analysis, we proposed PICA to improve black-box attacks. The motivation of PICA can be summarized as the two following aspects.

\textbf{1) It is sufficient to impose adversarial perturbations to a few key pixels.} Most of the existing attacks attempt to perturb the entire image, regardless of its spatial semantic information. In fact, imposing perturbations to the salient region may be easier to fool DNNs than the background \cite{dong2020robust}. This fact motivates us to restrict the perturbed pixels via the attentional mechanism. On the other hand, the one-pixel attack \cite{su2019one} has demonstrated that attacking a few key pixels might reach the goal in some cases. It also inspires us to concentrate on attacking limited pixels rather than a large number of pixels. Accordingly, we try to minimize the redundant attacks by considering the spatial semantic information and structural prior knowledge of the images.

\textbf{2) Existing EA-based attacks are less efficient for high-resolution images.} Recent evidence shows that the generation of AEs can be formulated as a MOP. A group of researchers has investigated the black-box attacks based on MOEAs, but the performance of current approaches is sensitive to the scale of target images. For high-resolution images, the dimension of decision variables can reach tens of thousands or more. Such a situation increases the difficulty in convergence with using conventional MOEAs. The generated perturbations will be much easier to perceptible (see Fig. \ref{1a}). These issues motivate us to consider to convert black-box adversarial attacks into low-dimensional MOPs for better finding the optimal perturbations.

\section{Methodology}
Fig. \ref{fig:framework} provides the overall framework of the proposed PICA. It mainly consists of three components, namely, 1) screening perturbed pixels with the attention map of the image; 2) refining perturbed pixels with the pixel correlation; 3)  generating perturbations with a general MOEA. In the following, we will detail each component step by step.

\subsection{Screening perturbed pixels with the attention mechanism}
\begin{figure}[htbp]
\centering
\includegraphics[width=3.2in]{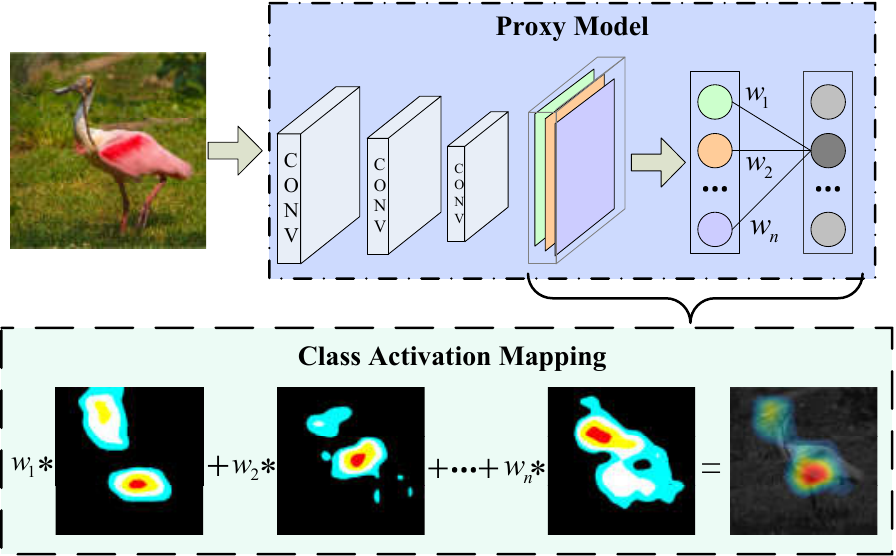}
\caption{The process of obtaining CAM with a proxy model.}
\label{fig:3}
\end{figure}

By considering the spatial semantic information, PICA first employs the class activation mapping (CAM) technique \cite{zhou2016learning} to access the attention map of the target image. CAM can visualize the scores of the predicted class of an image and highlight the pixels that significantly relate to the category. The attention map obtained with CAM can reflect the pixels of the interests of DNNs when recognizing a given image (as shown in Fig. \ref{fig:3}). Obviously, attacking these pixels that contain sufficient spatial semantic information can fool the target DNN with a relatively higher probability.  However, it is tricky to know the gradient information of black-box DNNs, which brings a barrier for using CAM. Under such an assumption, we suggest using a proxy model \footnote{We use SqueezeNet \cite{iandola2016squeezenet} as the proxy model in the experiments} to achieve an approximate attention map of the input image.

After that, we binarize the attention map generated with the proxy model to mark the candidate attacking pixels. The binarization is described as below.
\begin{equation}\label{bina}
{p_{i}} = \left\{
 \begin{array}{*{20}{c}}
 {0,{\rm{\quad}}{u_{i}} = 0}\\
 {1,{\rm{\quad}}{u_{i}} \ne 0}
 \end{array}\right.
\end{equation}
where $u_{i}$ is the pixel value at the $(l,w,c)$ position of the attention map (in detail, $l$ and $w$ determine the coordinate position of the pixel, and $c$ denotes the channel the pixel belongs to).  According to Eq. (\ref{bina}), the pixels with $p_{i} = 1$ are first screened as the candidate attacking pixels of a target image.

\subsection{Refining perturbed pixels with the pixel correlation}
The success of the one-pixel attack \cite{su2019one}, which refers to adding the perturbation to a selected pixel rather than the entire image,  inspires us to refine the screened perturbed pixels. Generally, the neighboring pixels of an image often share similar characteristics and pixel values \cite{he2015spatial}. We term this feature as the pixel correlation in this paper. Attacking all of them may lead to redundant perturbations, and more seriously, blur the visual feature of the target image. Thus, it is valuable to take advantage of the pixel correlation for determining the final perturbed pixels. A natural idea is to separate the image and extract a few pixels from each subcomponent, which sounds like the max-pooling. This implementation can result in a new image that is visually close to the original one. To minimize the perturbed pixels while maintaining the information of the original image as possible, we separate the image by parity pixels.  More concretely, we select only one pixel of every two neighboring ones as a candidate for adversarial attacks (see Fig. \ref{fig:framework}).

Mathematically, assume that the original image $I$ is with a size of $m\times n$, and the pixel values are all within a range of $[0,255]$. For refining the perturbed pixels, $I$ is divided into two parts, namely $I_1$, $I_2$, in the following way. For every $l$ and $w$, there exists
\begin{equation}\label{MOPd}
	\begin{array}{l}
		I_{1}^{(l, w)}=I^{(2l-1,2 w-1)}+I^{(2l, 2w)} \\
		I_{2}^{(l, w)}=I^{(2l-1,2 w)}+I^{(2l, 2w-1)}
	\end{array}
\end{equation}
where $1 \leq l \leq  [\frac{m}{2}]$ and $1 \leq w \leq [\frac{n}{2}]$. Note that, the new images $I_1$ and $I_2$ are visually indistinguishable compared with $I$, even though their sizes become half of the original size. Combing with the binarized attention map, the final perturbed pixels are determined according to Fig. \ref{fig:framework}. To summarize, the number of final perturbed pixels is less than that of the initial ones. Through leveraging the attentional mechanism and pixel-correlation, the dimension of decision variables greatly reduces, which facilitates the convergence of the subsequent black-box optimization-based attack.

\begin{figure*}[htbp]
	\centering
	\includegraphics[width=6.5in]{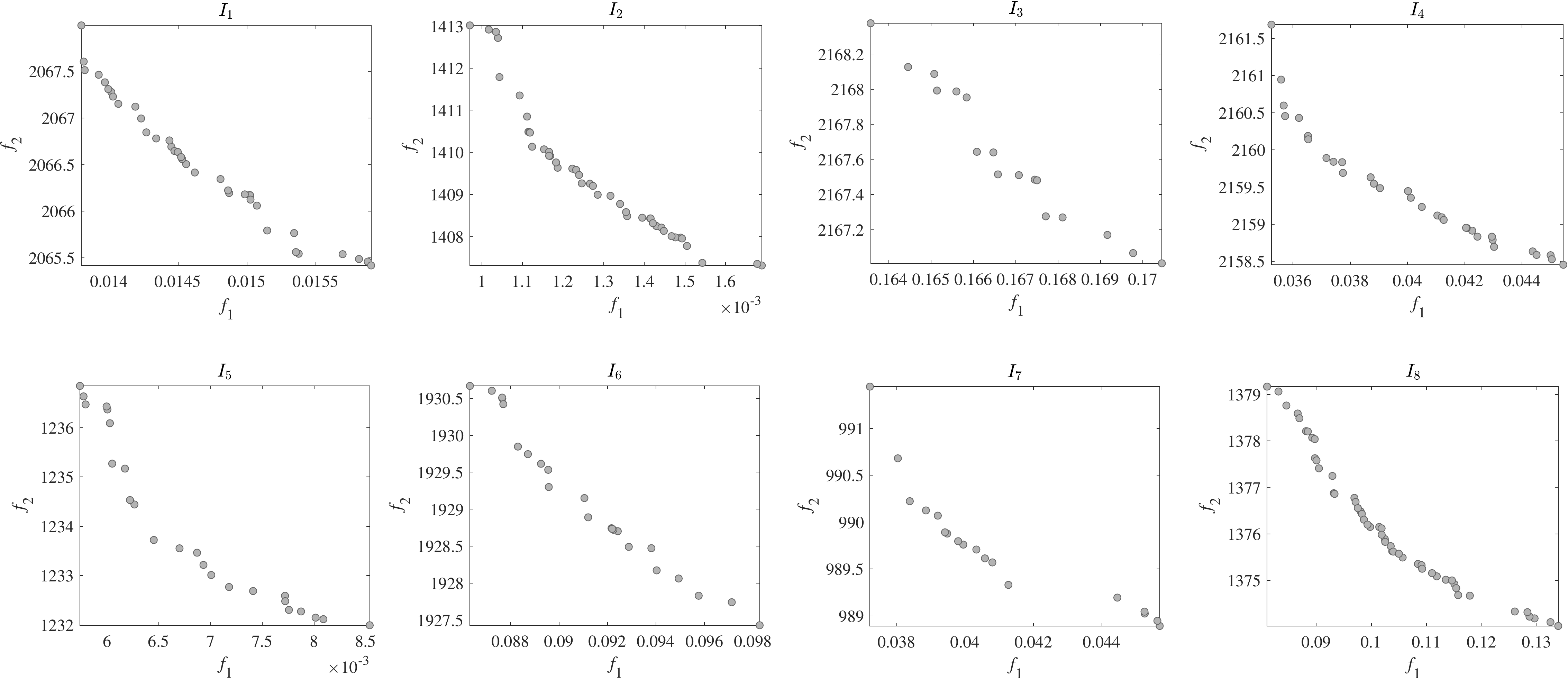}
	\caption{The Pareto optimal solutions obtained by PICA with eight randomly selected images.}
	\label{fig:pareto}
\end{figure*}

\begin{algorithm}[htbp]
	\caption{\emph{NSGA-II}}
	\begin{algorithmic}[1]\label{nsga}
		\REQUIRE $N$ (population size)
		\ENSURE $P$ (final population)
		\STATE $P \leftarrow$ \emph{Initilization}($N$);
		\WHILE{termination critertion not fullfilled}
		\STATE $P^{\prime} \leftarrow$ \emph{MatingSelection}($P$);
		\STATE $O \leftarrow$ \emph{GA}($P^{\prime}$);
		\STATE $P \leftarrow$ \emph{EnvironmentalSelection}($P\cup O$);
		\ENDWHILE
		\RETURN $P$
	\end{algorithmic}
\end{algorithm}	

\subsection{Generating perturbations with the MOEA}
In PICA, the black-box adversarial attack is formulated as a MOP below.
\begin{equation}\label{MOP}
	\begin{array}{cl}
		\min  & {f_1} = P(\mathcal{C}(\bm{I} + \bm{X} ) = \mathcal{C}(\bm{I}))\\
		\min  & {f_2} = {\left\| \bm{X}  \right\|_2}\\
		\mbox{s.t.} & 0 \le {u_{i}} + {x _{i}} \le 255
	\end{array}
\end{equation}
where $P(\cdot)$ denotes the \emph{confidence probability} of the classification result; $\bm{I}$ and $\bm{X}$ represent the \emph{original sample} and  \emph{adversarial perturbation}, respectively; $u_{i}$ is the value of pixel at the $(l,w,c)$ position of $\bm{I}$, while $x_{i}$ is the value of perturbation.

Seen from Eq. (\ref{MOP}), three remarks are given as follows. Firstly, the formulated MOP involves two conflicting objectives. The first one ${f_1}$ denotes the probability that the target DNN $\mathcal{C}( \cdot )$ classifies the generated AE $\bm{I} + \bm{X}$ into the correct class $\mathcal{C}(\bm{I})$. The second one represents the ${l_2}$ distance that is used to evaluate the similarity between $\bm{I} + \bm{X}$  and $\bm{I}$. More concretely, minimizing ${l_2}$ distance can reduce the candidate perturbation of each attacked pixel. Secondly, we note that most of the decision variables in $\bm{X}$ are fixed as zero, and the dimension of perturbations to be optimized is relatively lower than that of $\bm{I}$. The reason is that the candidate attacked pixels are significantly reduced according to the techniques introduced in the last two subsections. As a consequence, the formulated MOP is available for most of the existing MOEAs without involving any techniques for large-scale MOPs \cite{qian2017so}. Thirdly, the constraint imposed to $\bm{X}$ defines the range of perturbation on each attacked pixel. Using such prior knowledge can limit the search space and facilitate the convergence of MOEA.

The proposed PICA is compatible with any MOEA. Taking NSGA-II \cite{deb2002fast} as an example, we describe the pseudo-code of the embedded optimizer in Algorithm \ref{nsga}. To begin with, the population $P$ is initialized, that is, a set of perturbed images are generated. After the initialization, $N$ parents are first selected based on their nondominated front numbers and crowding distances through the binary tournament selection. Then, the offspring set $O$ is generated by applying simulated binary crossover (SBX) \cite{deb2011multi} and polynomial mutation (PM) \cite{deb1996combined} on $P^{\prime}$. Finally, $N$ perturbed images with better nondominated front numbers and crowding distances survive from the combination of $O$ and $P^{\prime}$. The procedure is terminated until the number of function evaluations reaches the maximum number.

\begin{table*}[htbp]
	\centering
\small
	\renewcommand{\arraystretch}{1.4}
	\begin{tabular}{c|cc|cc|cc|cc}
		\hline
       \multirow{3}{*}{Image No.}&   \multicolumn{8}{c}{Recognition results and confidence}\\ \cline{2-9}
		&   \multicolumn{4}{c|}{ResNet-101}&   \multicolumn{4}{c}{Inception-v3}\\ \cline{2-9}
		& \multicolumn{2}{c|}{$\mathcal{C}(\bm{I})$}&    \multicolumn{2}{c|}{$\mathcal{C}(\bm{I} + \bm{X})$}&   \multicolumn{2}{c|}{$\mathcal{C}(\bm{I})$}&  \multicolumn{2}{c}{$\mathcal{C}(\bm{I} + \bm{X})$}\\\hline
		$I_1$&         Teapot:&                30.71\%&         Bakery:&           30.34\%&     Teapot:&          37.22\%&     Soap dispenser:&  88.96\% \\
		$I_2$&         Bottlecap:&             98.78\%&         Baseball:&         99.71\%&     Bottlecap:&       74.13\%&     Baseball:&        98.56\% \\
		$I_3$&         Daisy:&                 78.49\%&         Vase:&             34.68\%&     Ant:&             28.38\%&     Ant:&             97.86\% \\
		$I_4$&         Pencil box:&            46.79\%&         Pencil box:&       92.02\%&     Pencil box:&      39.12\%&     Pencil box:&      64.10\%\\
		$I_5$&         Maltese dog:&           66.44\%&         French bulldog:&   13.01\%&     Maltese dog:&     93.32\%&     Shih-Tzu:&        35.29\%\\
		$I_6$&         Acorn:&                 99.49\%&         Goldfinch:&        22.20\%&     Acorn:&           95.21\%&     Fig:&             46.32\%\\
		$I_7$&         Beacon:&                91.24\%&         Breakwater:&       68.14\%&     Beacon:&          89.05\%&     Breakwater:&      30.78\%\\
		$I_8$&         Flamingo:&              99.70\%&         Crane:&            86.06\%&     Flamingo:&        86.78\%&     White stork:&     75.75\%\\ \hline
	\end{tabular}
    \vspace{1mm}
	\caption{Classification results and corresponding confidences of the original and AEs.}	\label{tab:1}
\end{table*}
\begin{table}[htbp]
	\centering
	\footnotesize
	\renewcommand{\arraystretch}{1.6}
	\begin{tabular}{c|c|c}
		\hline
		Target model&                       Attack method&                Classification accuracy\\\hline
		\multirow{3}{*}{ResNet-101}&      N/A&                   85.00\% \\ \cline{2-3}
		&      with baseline method&  7.00\% \\\cline{2-3}
		&      with PICA&      0.00\% \\\hline
		\multirow{3}{*}{Inception-v3}&    N/A&       83.00\% \\ \cline{2-3}
		&      with baseline method&  18.00\% \\\cline{2-3}
		&      with PICA&      2.00\% \\\hline
		
	\end{tabular}
\vspace{1mm}
	\caption{Classification results and corresponding confidences of the original and AEs.}
	\label{tab:2}
\end{table}

\begin{figure*}[htbp]
	\vspace{8mm}
	\centering
	\subfigure[Original samples]{
		\includegraphics[width=6.8in]{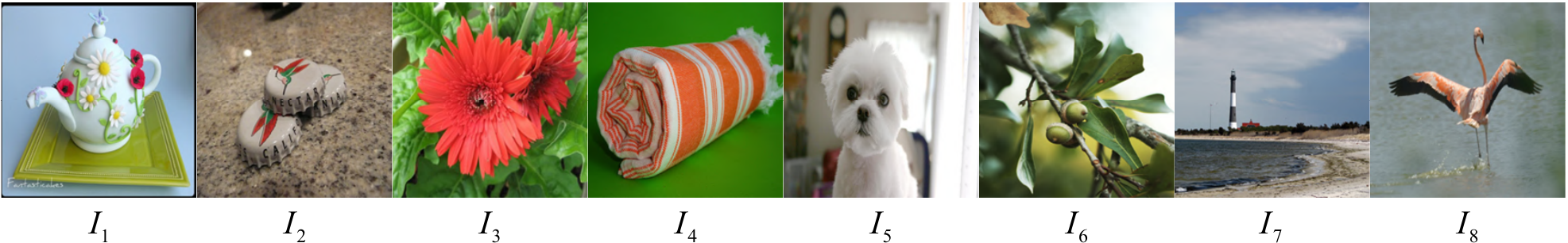}
	}
	\subfigure[Perturbation patterns and AEs 
	on ResNet-101]{
		\includegraphics[width=6.8in]{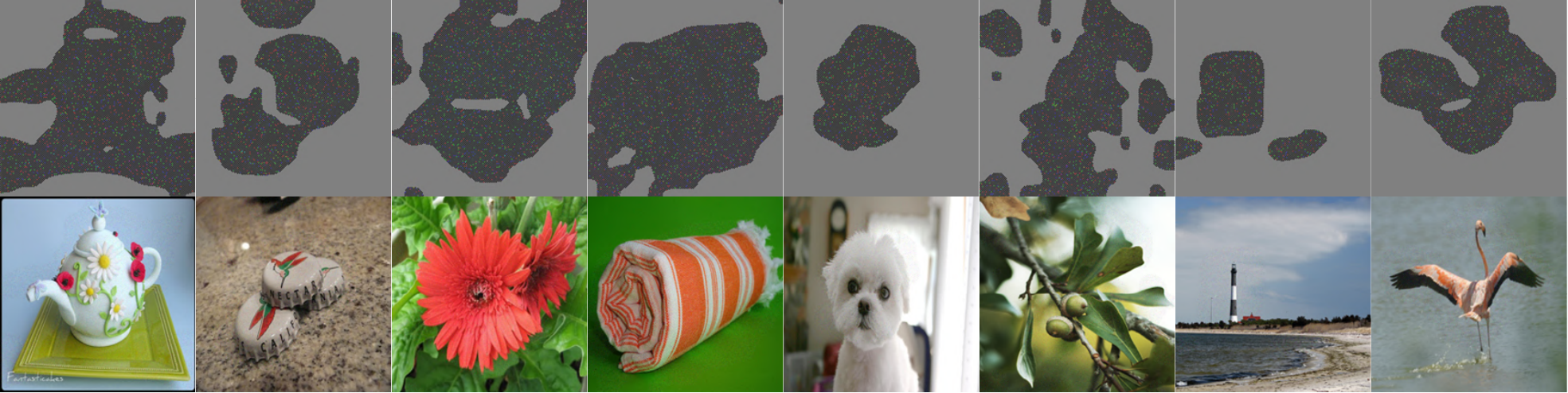}
	}
    \subfigure[Perturbation patterns and AEs
     on Inception-v3]{
		\includegraphics[width=6.8in]{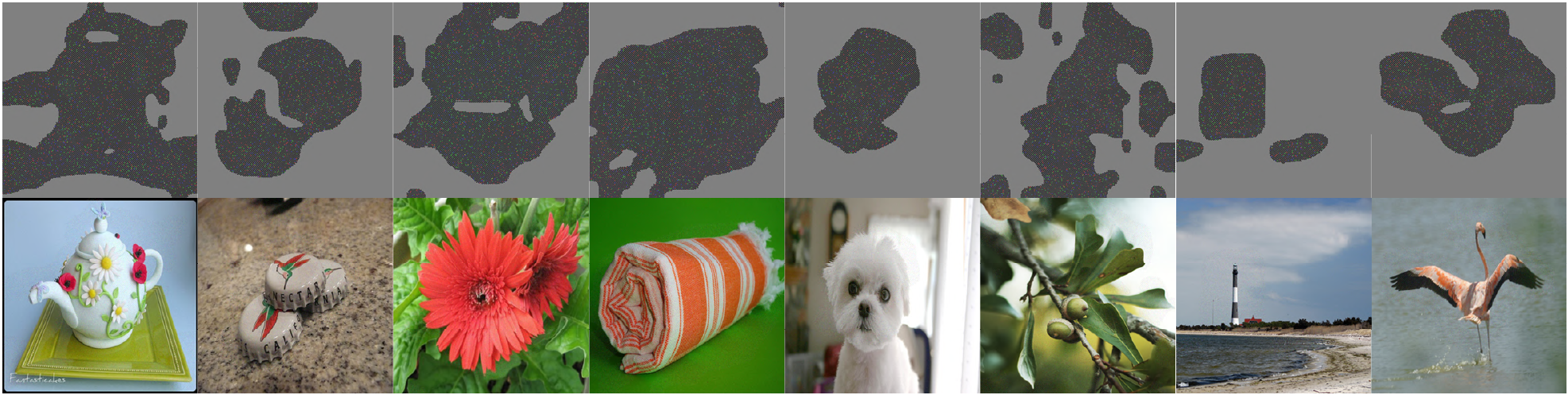}
	}

\caption{Illustration of AEs obtained by the proposed PICA with different target DNNs.}
	\label{fig:4}
\end{figure*}

\section{Experiments}
In this section, we first introduce the experimental setup, including the benchmark dataset and parameter setting. Then, we investigate the performance of the proposed PICA and compare it with an MOEA-based adversarial attack. Finally, the ablation study is performed to verify the effectiveness of using the attentional mechanism and pixel correlation.

\subsection{Experimental Setup}
The benchmark dataset used in this paper consists of $100$ high-resolution images randomly chosen from ImageNet-1000 \cite{deng2009imagenet}.  The pretrained ResNet-101 \cite{he2016deep} and Inception-v3 \cite{szegedy2016rethinking} are specified as the target models for the study on the robustness of the proposed PICA, owing to their different scales of the input layers. 

As for the parameter setting in MOEA,  the population size is set to $50$ for each target image. The number of maximum function evaluations is set to $10,000$. The distribution indexes of SBX and PM are fixed as 20. The probabilities of crossover and mutation are fixed as $1.0$ and $1/d$, where $d$ is the dimension of the decision variables. All the experiments are carried out on a PC with Intel Core i7-6700K 4.0GHz CPU, 48GB RAM, Windows 10, and Matlab R2018b with PlatEMO \cite{tian2017platemo}.

\subsection{Results and Analysis}
Fig. \ref{fig:pareto} exhibits the Pareto optimal solutions obtained with the proposed PICA on eight randomly selected images in the benchmark dataset. We can observe that the proposed PICA has reached a trade-off between the two conflicting objectives in each subfigure. Closer inspection reveals that the solutions with high confidence of the true category usually have a low $l_{2}$ norm, and vice versa. It is also worth noting that each solution corresponds to a perturbed image, and only a few of these images can be defined as AEs. Without loss of generality, we identify the perturbed image that successfully fools the target DNN with the minimum $l_{2}$ norm as the final AE.

\begin{figure}[htbp]
	\centering
	\subfigure[AEs generated by the baseline method]{
		\includegraphics[width=3.1in]{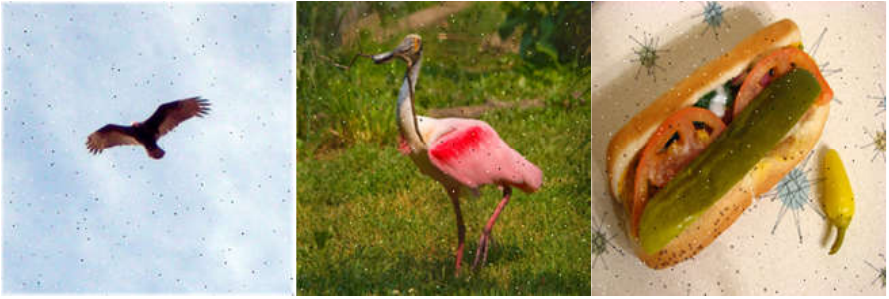}
	}
	\subfigure[AEs generated by the PICA]{
		\includegraphics[width=3.1in]{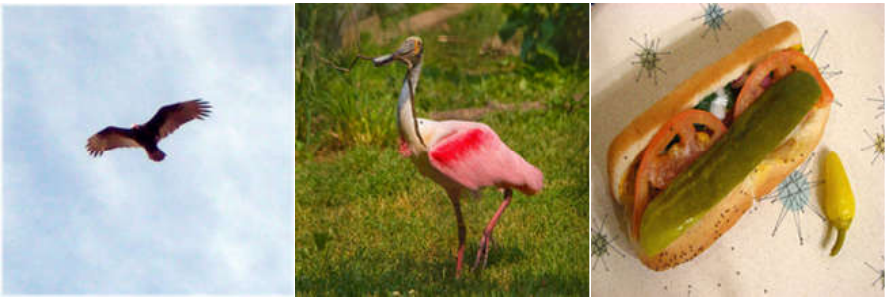}
	}
	\caption{The visual comparison of the AEs generated by the baseline method and the PICA.}
	\label{fig:compare}
\end{figure}

Table \ref{tab:1} shows the classification results and confidence probabilities of the above eight original images and their corresponding AEs obtained with ResNet-101 and Inception-v3. Two remarks can be summarized as follows from the table. First of all, the proposed PICA can generate AE with high confidence probabilities (at least $22.20\%$, most of the results over $60\%$), although most of the original images are correctly classified by the two DNNs. Secondly, for the misclassified images on the two DNNs ($I_{4}$ for ResNet-101; $I_{3}$ and $I_{4}$ for Inception-v3), the proposed PICA further improves the confidence probabilities of the incorrect category (from $46.79\%$ to $92.02\%$ on $I_{4}$ for ResNet-101, $28.38\%$ to $97.86\%$ on $I_{3}$ for Inception-v3, and $39.12\%$ to $64.10\%$ on $I_{4}$ for Inception-v3). We also find that the classification results of AEs are quite different from those of the original images. For example, the generated AE of $I_{6}$ is recognized as an animal, whereas the original category belongs to a plant. The above observation illustrates that the proposed PICA can significantly mislead DNNs rather than resulting in AEs with similar categories compared with the original ones.

Fig. \ref{fig:4} shows eight AEs obtained by the proposed PICA against the two DNNs. It is noteworthy that the value of perturbations includes both positive and negative values. For better visualization, the unperturbed pixels in each image are exhibited in dark gray, whereas the pixels that reside out of the salient region are in light gray. More importantly, the brighter pixels indicate the increases of intensity, while darker pixels represent the intensity changes in the opposite direction. As shown in Fig. \ref{fig:4}, the AEs generated by PICA have similar visual perception compared with the original images, regardless of the structure of target DNNs. The reason is that the overall intensity of the perturbations is relatively low, and the changes only emerge in a few pixels of images. Therefore, the above evidence verifies the effectiveness of the proposed PICA.

\begin{figure}[htbp]
	\centering
	\subfigure[Baseline method]{
		\includegraphics[width=3.1in]{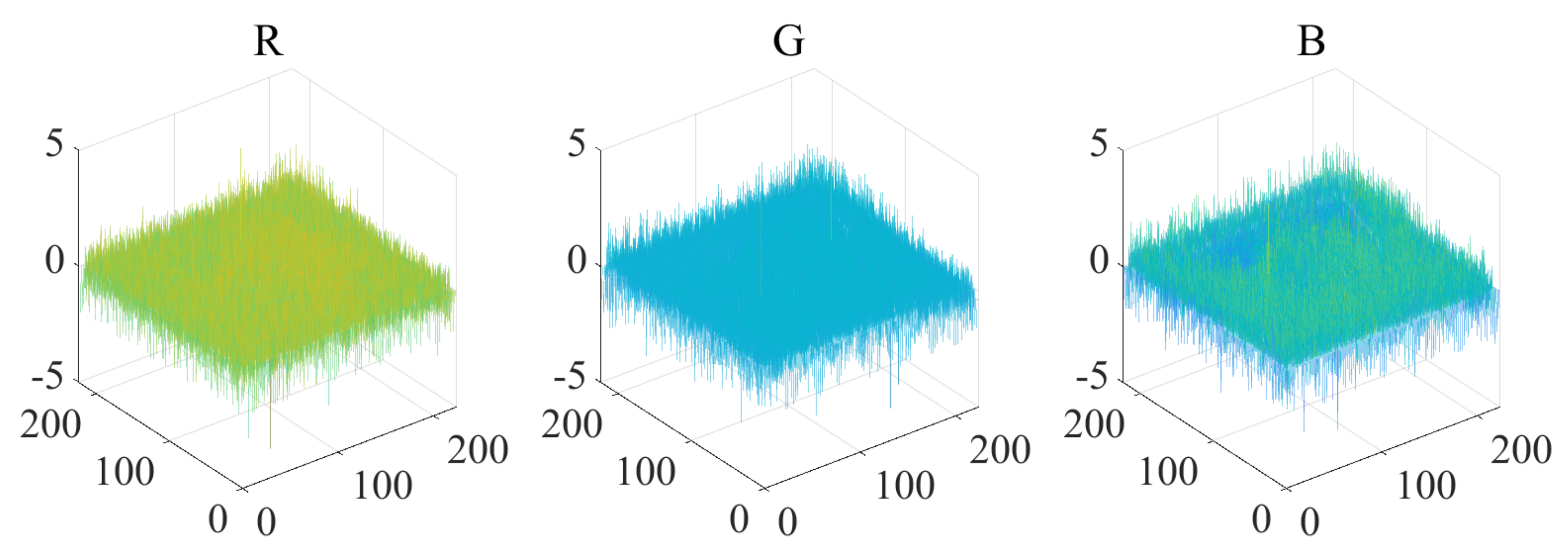}
	}
	\subfigure[PICA]{
		\includegraphics[width=3.1in]{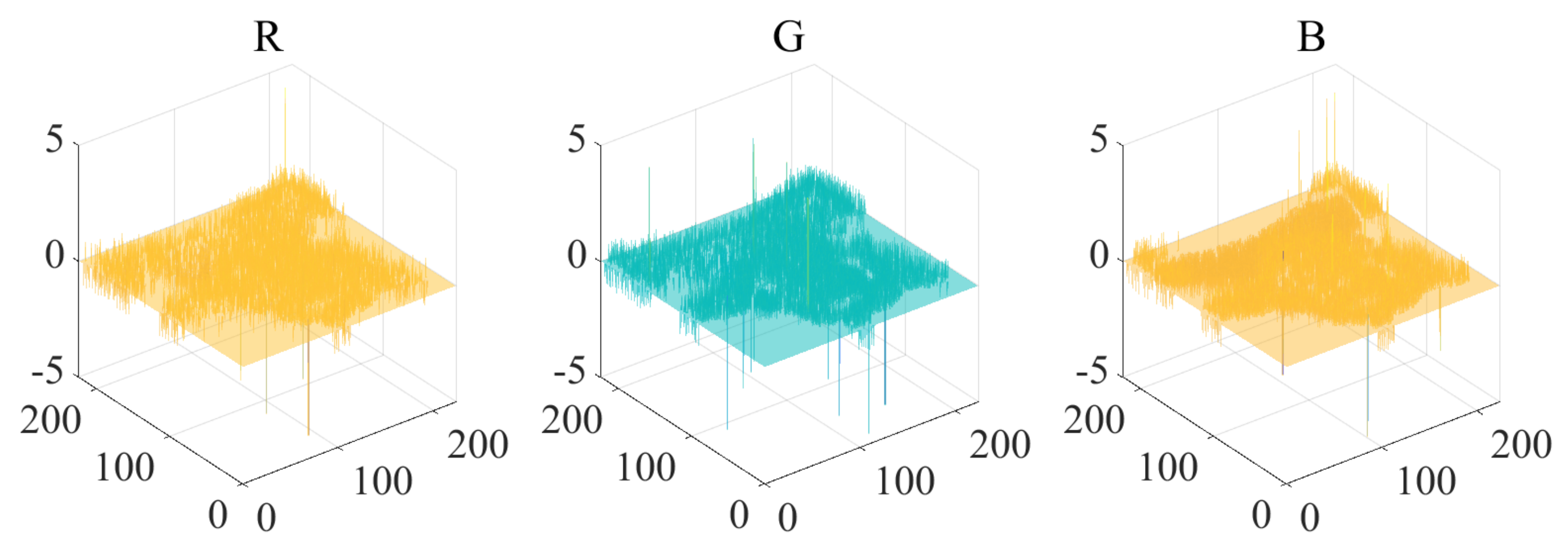}
	}
	\caption{Illustration of the attacking results without and with the attentional mechanism.}
	\label{fig:attention}
\end{figure}

Furthermore, we investigate the performance comparison of PICA and the baseline method suggested in \cite{suzuki2019adversarial}, which follows the block-division pattern and solves the black-box attack with MOEA/D \cite{zhang2007moea}. Table \ref{tab:2} statistically compares them on both two DNNs. From the table, we can observe that 85\% of the benchmark images can be correctly classified by ResNet-101, while 17\% of them are misclassified by Inception-v3. After performing the attack with the baseline method, only 7\% and 18\% of the images are correctly classified by the two DNNs, respectively. By contrast, PICA has successfully attacked most of the images and fooled both of the two models. More concretely, PICA has achieved a 100\% success rate on ResNet-101. Figure \ref{fig:compare} further visualizes the AEs generated by the two methods. As for the baseline method, some of its AEs have much visible noise, which is easily captured by human vision. For PICA, the perturbations are more difficult to perceive. This is attributed to the usage of attentional mechanism and pixel correlation, which refines the perturbed pixels and benefits for the search of the MOEA.

\subsection{Ablation Study}
\subsubsection{Effectiveness of using attentional mechanism}
First of all, we investigate the influence of the attentional mechanism on the proposed PICA. In detail, we randomly select an image from the benchmark dataset and perform MOEA-based black-box attacks subject to two settings below. The first one is attacking the screened pixels with only considering the pixel correlation. The second one is perturbing the refined pixels by using both attentional map and pixel correlation, that is, performing PICA on the selected image.

\begin{figure}[htbp]
	\centering
	\subfigure[Searching only with attentional mechanism]{
		\includegraphics[width=3.1in]{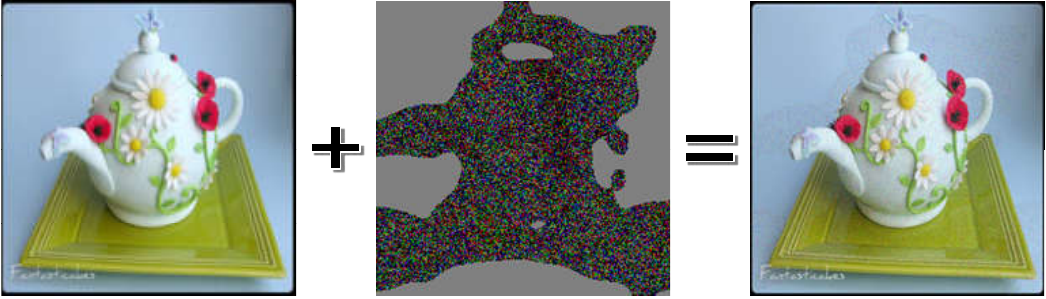}
	}
	\subfigure[Searching with attentional mechanism and  pixel correlation]{
		\includegraphics[width=3.1in]{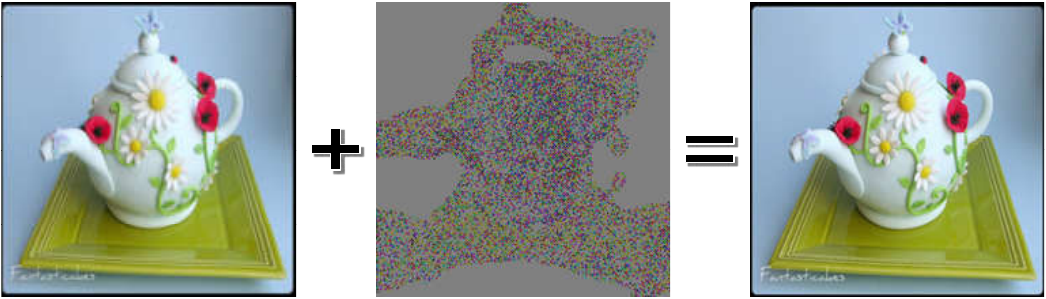}
	}
	\caption{Illustration of the attacking results without and with considering pixel correlation, respectively.}
	\label{fig:template}
\end{figure}

Fig. \ref{fig:attention} provides the attacking results under two different settings, where the perturbations in each channel of two AEs are visualized from top to bottom, respectively. From the observation, we can find that the bottom AE has less number of perturbed pixels compared with the top one. Besides, the less perturbed pixels do not bring an unacceptable intensity of perturbations for the bottom AE. The fact highlights the advantage of using spatial semantic information. It also indicates that PICA can obtain invisible AEs while ensuring the success rate of attacking with the help of the attentional mechanism, although ignoring a large number of pixels.

\subsubsection{Effectiveness of using pixel correlation}
Other than the attentional mechanism, we second investigate the effectiveness of using pixel correlation in the proposed PICA. Likewise, we also randomly select an image and perform attacks with PICA and its variant. To be specific, the variant of PICA does not take the pixel correlation into account.

Fig. \ref{fig:template} compares the generated AEs with the attentional mechanism only and with both two techniques. Although the variant allows for a successful attack, the intensity of perturbation generated is still larger than PICA. It is also clear to see that the bottom AE in Fig. \ref{fig:template} is much more imperceptible. The above results also confirm the validity of PICA, which refines the attacked pixels based on the pixel correlation.

\section{Conclusion}
This study has proposed a novel MOEA-based black-box adversarial attack, termed as PICA. It has considered the spatial semantic information and pixel-correlation of images to restrict the target pixels. These techniques used for dimension reduction are proven to facilitate the attack via MOEAs. Experimental results have shown that PICA can effectively perform black-box adversarial attacks while ensuring the visual quality of candidate AEs. In comparison with the state-of-the-art methods, PICA also highlights its superiority and potential on high-resolution images, such as the ImageNet dataset. Our future works will include investigating a proxyless pixel reduction and designing a specialized MOEA for black-box adversarial attack.

{\small
\bibliographystyle{ieee_fullname}
\bibliography{cvpr}
}

\end{document}